\documentclass[twocolumn, 10pt]{article}
\usepackage{graphicx} 

\usepackage{cite}
\usepackage{amsmath,amssymb,amsfonts}
\usepackage{algorithmic}
\usepackage{graphicx}
\usepackage{textcomp}
\usepackage{xcolor}
\usepackage{float}
\usepackage{comment}
\usepackage{enumitem}
\usepackage{amssymb}
\usepackage{mathrsfs}
\usepackage{amsmath}
\usepackage{geometry}
\usepackage{setspace}
\usepackage{enumitem}
\usepackage{amssymb}
\usepackage{setspace}
\usepackage{adjustbox}
\usepackage{float}
\usepackage{booktabs}
\usepackage{amsmath}
\usepackage{xcolor}
\usepackage{acronym}
\usepackage{url}
\usepackage{relsize}

\usepackage[pdfborder={0 0 0}, colorlinks=true, linkcolor=black, citecolor=black, urlcolor= black]{hyperref}
\def\BibTeX{{\rm B\kern-.05em{\sc i\kern-.025em b}\kern-.08em
    T\kern-.1667em\lower.7ex\hbox{E}\kern-.125emX}}

\makeatletter
\renewenvironment{abstract}{%
  \par\small
  \noindent\textbf{Abstract—}\ignorespaces
}{%
  \par\normalsize\bfseries
}
\makeatother

\usepackage{tikz}

\newcommand{\copyrightnotice}{
\begin{tikzpicture}[remember picture,overlay]
\node[anchor=south,yshift=10pt] at (current page.south) {
\fbox{\parbox{\dimexpr\textwidth-2\fboxsep-2\fboxrule\relax}{
\footnotesize
© 2024 IEEE. Personal use of this material is permitted. Permission from IEEE must be obtained for all other uses, in any current or future media, including reprinting/republishing this material for advertising or promotional purposes, creating new collective works, for resale or redistribution to servers or lists, or reuse of any copyrighted component of this work in other works.

This is the author’s version of a paper accepted for publication in the \href{https://doi.org/10.1109/ICCTA64612.2024}{IEEE 34th International Conference on Computer Theory and Applications (ICCTA)}, 2024. The published version is available via DOI: \href{https://doi.org/10.1109/ICCTA64612.2024.10974880}{10.1109/ICCTA64612.2024.10974880}.
}}
};
\end{tikzpicture}
}

\usepackage{titlesec}

\renewcommand{\thesection}{\Roman{section}}
\renewcommand{\thesubsection}{\Alph{subsection}}
\renewcommand{\thesubsubsection}{\arabic{subsubsection}}

\titleformat{\section}
  {\normalfont\filcenter\MakeUppercase}
  {\thesection.}
  {0.6em}
  {}

\titleformat{\subsection}
  {\normalfont\itshape}
  {\thesubsection.}
  {0.6em}
  {}

\titleformat{\subsubsection}[runin]
  {\normalfont\itshape}
  {\thesubsubsection)}
  {0.6em}
  {}
  [.\ ] 

\titlespacing*{\section}{0pt}{1.2ex plus 0.3ex minus 0.2ex}{0.8ex}
\titlespacing*{\subsection}{0pt}{1.0ex plus 0.3ex minus 0.2ex}{0.6ex}
\titlespacing*{\subsubsection}{0pt}{0.8ex plus 0.2ex minus 0.2ex}{0.6em}

\begin{document}

\makeatletter
\renewcommand{\maketitle}{%
  \twocolumn[{%
    \begin{@twocolumnfalse}
      \vspace*{-1.2em}
      \begin{center}
        {\huge\bfseries \@title \par} 
        \vspace{0.8em}
        {\normalsize
        \begin{tabular}{@{}ccc@{}}
          \begin{tabular}{@{}c@{}}
            \textbf{Youssef Mahran}\\
            \textit{Mechatronics Engineering Department}\\
            \textit{The German University in Cairo}\\
            Cairo, Egypt\\
            \texttt{youssef.mahran@student.guc.edu.eg}
          \end{tabular}
          &
          \begin{tabular}{@{}c@{}}
            \textbf{Zeyad Gamal}\\
            \textit{Mechatronics Engineering Department}\\
            \textit{The German University in Cairo}\\
            Cairo, Egypt\\
            \texttt{zeyad.abdrabo@student.guc.edu.eg}
          \end{tabular}
          &
          \begin{tabular}{@{}c@{}}
            \textbf{Ayman El-Badawy}\\
            \textit{Mechatronics Engineering Department}\\
            \textit{The German University in Cairo}\\
            Cairo, Egypt\\
            \texttt{ayman.elbadawy@guc.edu.eg}
          \end{tabular}
        \end{tabular}\par
        }
      \end{center}
      \vspace{0.8em}
    \end{@twocolumnfalse}
  }]%
}
\makeatother

\title{Dynamic Entropy Tuning in Reinforcement Learning Low-Level Quadcopter Control:\\Stochasticity vs Determinism}

\date{}

\maketitle
\copyrightnotice
\begin{abstract}
\bfseries This paper explores the impact of dynamic entropy tuning in Reinforcement Learning (RL) algorithms that train a stochastic policy. Its performance is compared against algorithms that train a deterministic one. Stochastic policies optimize a probability distribution over actions to maximize rewards, while deterministic policies select a single deterministic action per state. The effect of training a stochastic policy with both static entropy and dynamic entropy and then executing deterministic actions to control the quadcopter is explored. It is then compared against training a deterministic policy and executing deterministic actions. For the purpose of this research, the Soft Actor-Critic (SAC) algorithm was chosen for the stochastic algorithm while the Twin Delayed Deep Deterministic Policy Gradient (TD3) was chosen for the deterministic algorithm. The training and simulation results show the positive effect the dynamic entropy tuning has on controlling the quadcopter by preventing catastrophic forgetting and improving exploration efficiency.
\end{abstract}

\section{Introduction}
Recent advancements in RL showed huge potential in controlling quadcopters. This was proven by allowing a quadcopter to obtain champion-level drone racing performance \cite{nature}. The quadcopter was able to navigate the race course and pass through all the gates in the correct sequence faster than the human-controlled quadcopter and win against 3 different drone racing champions achieving the fastest recorded race time. This shows the potential of using the RL control framework in controlling quadcopters.

Deterministic RL algorithms were explored in the literature to better study the control of quadcopters. The TD3 algorithm is a powerful deterministic algorithm with its dual-clipped Q-learning and delayed policy update techniques. Two different RL agents were developed to handle two different quadcopter control tasks using TD3 \cite{mazen}. The main goal of the first agent, known as the stabilization agent, was to maintain the quadcopter hovering at a certain pre-set point, while the second agent, known as the path-following agent, was to navigate the quadcopter along a predetermined path. Both agents mapped environment states into motor commands. Simulation results showcased the efficiency of both agents. To further enhance the quadcopter capabilities, an obstacle avoidance agent (OA) was created in addition to the path following agent (PF) \cite{mokhtar2023autonomous}. The target of the PF agent was to map the states into motor RPMs to follow a certain path. The OA agent modified the tracking error information before it was sent to the PF agent to ensure that the path was free from any obstacles. Simulation results proved that this control framework was successful in following paths while avoiding obstacles along the way. Another control framework in the form of high-level and low-level waypoint navigation of the quadcopter was proposed \cite{HIMANSHU2022281}. The low-level controller maps the environment states into motor commands while the high-level controller produces the linear velocities of the quadcopter. Both agents' simulation results showed the successful navigation and trajectory tracking of the quadcopter.

The SAC algorithm on the other hand combines two powerful reinforcement learning techniques, the Actor-Critic technique and maximum entropy deep RL. Combining these two techniques makes the SAC a state-of-the-art algorithm. The SAC differs from the TD3 by adding an entropy term to the target and the policy. This technique helps maximize the exploration while taking into consideration maximizing the rewards as well. It also computes the next-state action from the current policy and does not utilize target policy smoothing. This is due to the SAC algorithm training a stochastic policy and the stochastic noise is enough to smooth the Q-values. The SAC algorithm was explored in the literature by proposing a SAC low-level control of a quadcopter that maps environment states to motor commands \cite{DBLP:journals/corr/abs-2010-02293}. The simulation results show the efficiency of the controller in a simple go-to task and in tracking moving objects at high and low speeds with maximum efficiency. The agent was then tested with various extreme initial positions to test its robustness and the SAC algorithm's ability in state space exploration. However, the proposed algorithm utilized static entropy only. Another  SAC agent for a quadcopter was developed to control the quadcopter despite having a single rotor failure \cite{DBLP:journals/corr/abs-2109-10488}. Simulation results showed the SAC's effectiveness in hovering, landing, and tracking various trajectories with only three active rotors. The controller was also successfully tested against wind disturbances to validate its robustness. Also in this research, the effect of dynamic entropy tuning was not explored.

A key factor in the SAC algorithm's performance is the entropy value. Entropy, in the context of reinforcement learning, refers to the randomness or unpredictability of the agent's action selection. Higher entropy encourages the agent to explore a wider range of actions, while lower entropy leads to more deterministic and exploitative behavior. Choosing the optimal entropy in entropy-based algorithms is a non-trivial task where the entropy is required to be tuned for each application \cite{haarnoja2019soft}. Dynamically adjusting the entropy allows the policy to balance exploration and exploitation effectively throughout the training process. The entropy is tuned by adjusting the entropy coefficient to match a target entropy value to minimize the difference between the current policy's entropy and the target entropy. This automatic tuning mechanism allows the agent to explore more when necessary and exploit more in explored regions, leading to more efficient and effective learning.

This paper aims to explore the impact of dynamic entropy tuning on low-level control of quadcopters. While deterministic algorithms are used more than stochastic algorithms for quadcopter control, this research compares stochastic and deterministic policy training. Both policies are tested within a deterministic environment to identify the optimal training approach. The comparison of these policy training methods and the impact of dynamic entropy tuning on quadcopter control has never been proposed in the literature before.

\section{Methodology}

\subsection{Algorithms Network Structure}
 TD3 and SAC algorithms were selected due to their effectiveness in continuous control tasks. The TD3 algorithm represents our deterministic algorithm while the SAC algorithm represents the stochastic one. The selected neural network architecture for all networks of the two algorithms consists of two hidden layers with 400 and 300 nodes each, one input layer, and an output layer with one output for the critic and four outputs for the actor. LeakyReLU activation functions are used in both hidden layers.
 
\subsection{Reward Function \& Entropy Tuning}
Entropy in the context of reinforcement learning refers to the unpredictability of the agent's action. The entropy is represented as $H(\pi)$ as shown in Eq. \ref{entropy}. Where ($\pi(a|s)$) is the policy of the expected action ($a$) in the given state ($s$).
\begin{equation}
    H(\pi) = - \log \pi(a|s)
    \label{entropy}
\end{equation}

 The goal of RL algorithms is to learn a policy ($\pi$) that maximizes the discounted sum of all possible future rewards as denoted in Eq. \ref{policy}. Where ($\gamma^t$) is the discount factor and $\big(r(t)\big)$ is the reward function shown in Eq. \ref{ec}.
\begin{equation}
    \pi^* = \substack{\mathlarger{\text{argmax}} \\ \mathlarger{\pi}} \hspace{2pt} \substack{\mathlarger{\text{E}} \\ \tau \thicksim \pi} \bigg[\sum_{t=0}^{\infty} \gamma^t \Big(r(t) \Big) \bigg]
    \label{policy}
\end{equation}

The reward function presented in Eq. \ref{reward} consists of two different parts. The first part is a rational function with $a = 7$ and $||\vec{e}_{k}||_{2}$ representing the Euclidean distance as shown in Eq. \ref{ec}. This part is responsible for tracking the stable hovering of the quadcopter. The second part is a normal distribution function with a standard deviation of 0.5 responsible for tracking the quadcopter with each step closer to the target yields a higher positive reward \cite{mokhtar2023autonomous}.

    \begin{equation}
    r(t) = \frac{1}{a \text{  }* \text{  }||\vec{e}_{k}||_{2}} + \frac{a}{\sqrt{2\pi \sigma^2}} e^{-0.5(\frac{||\vec{e}_{k}||_{2}}{\sigma})^2}
    \label{reward}
    \end{equation}
    \begin{equation}
    ||\vec{e}_{k}||_{2} = \sqrt{\Delta x^2 + \Delta y^2 + \Delta z^2}
    \label{ec}
    \end{equation}

However, learning an optimal policy decreases the entropy as each state ($s$) has an optimal action ($a$) decreasing the randomness in action selection. To solve this, an entropy bonus is added to the policy to encourage the agent to execute unpredictable actions while also maximizing the reward  \cite{haarnoja2018soft}. So, the policy function is changed as shown in Eq. \ref{entbon}. Where ($\alpha$) is the entropy coefficient and ($R(t, \pi)$ is the new reward function in Eq. \ref{rewent}. 
\begin{equation}
    \pi^* = \substack{\mathlarger{\text{argmax}} \\ \mathlarger{\pi}} \hspace{2pt} \substack{\mathlarger{\text{E}} \\ \tau \thicksim \pi} \bigg[\sum_{t=0}^{\infty} \gamma^t \Big(R(t, \pi) \Big)\bigg]
    \label{entbon}
\end{equation}

This changes our reward function presented in Eq. \ref{reward} to a reward function with entropy bonuses added as shown in Eq. \ref{rewent}
\begin{equation}
    R(t, \pi) = r(t) + \alpha H(\pi)
    \label{rewent}
    \end{equation}

The framework presented prevents an agent from quickly converging to a policy that is only locally optimal and may not be globally optimal. In hard exploration problems, the agent may settle for a locally optimal policy. Increasing the entropy means increasing the agent's adaptability. This framework is called the maximum entropy reinforcement learning as we are solving the traditional RL problem of maximizing the reward while jointly maximizing the entropy. This approach helps the agent generalize the policy and adapt to unseen circumstances. 

Adjusting the entropy coefficient is crucial in the goal of maximum entropy RL. Setting it to a static non-changing value is a poor solution since the agent should be able to explore in states that have not been explored yet, and exploit in already visited states \cite{haarnoja2019soft}. Dynamic entropy solves this problem by adjusting the entropy coefficient depending on the training needs while helping to achieve maximum entropy RL.

Instead of choosing an entropy value and fixing it, a minimum lower bound for the entropy is chosen as the target entropy. Typically this lower bound is equal to negative the dimension of the action space as shown in in Eq. \ref{bound}. Where ($H_{0}$) is the target entropy and ($A$) is the action space presented in Eq. \ref{action}. The bigger the action space the lower the entropy bound increasing the stochasticity of the policy. 
\begin{equation}
        H_{0} = - dim(A)
        \label{bound}
    \end{equation}
    
This turns the reinforcement learning problem shown in Eq. \ref{entbon} into a constrained one. Where the entropy is constrained to a lower bound ($H_{0}$). The entropy coefficient is then tuned to minimize the difference between the current policy's entropy ($H(\pi)$) and the target entropy ($H_{0}$). The entropy coefficient loss function is shown in Eq. \ref{loss}.
\begin{equation}
        J(\alpha) = \substack{\mathlarger{\text{E}} \\ s_{t} \thicksim D}\hspace{5pt} \Big[\hspace{2pt}\substack{\mathlarger{\text{E}} \\ a_{t} \thicksim \pi_{\vartheta}(.|s_{t})}\hspace{5pt} [-\alpha H(\pi) - \alpha H_{0}] \Big]
        \label{loss}
    \end{equation}

After calculating the loss function, gradient descent is then used to update the parameters of the entropy to minimize the error between the current policy's entropy and the target entropy. This framework of dynamically tuning the entropy, helps in maximizing the trade-off between expected reward and the entropy, balancing between exploration and exploitation.

 \subsection{Reinforcement Learning Framework}
\begin{figure}[H]
\centerline{\includegraphics[width=0.4\textwidth]{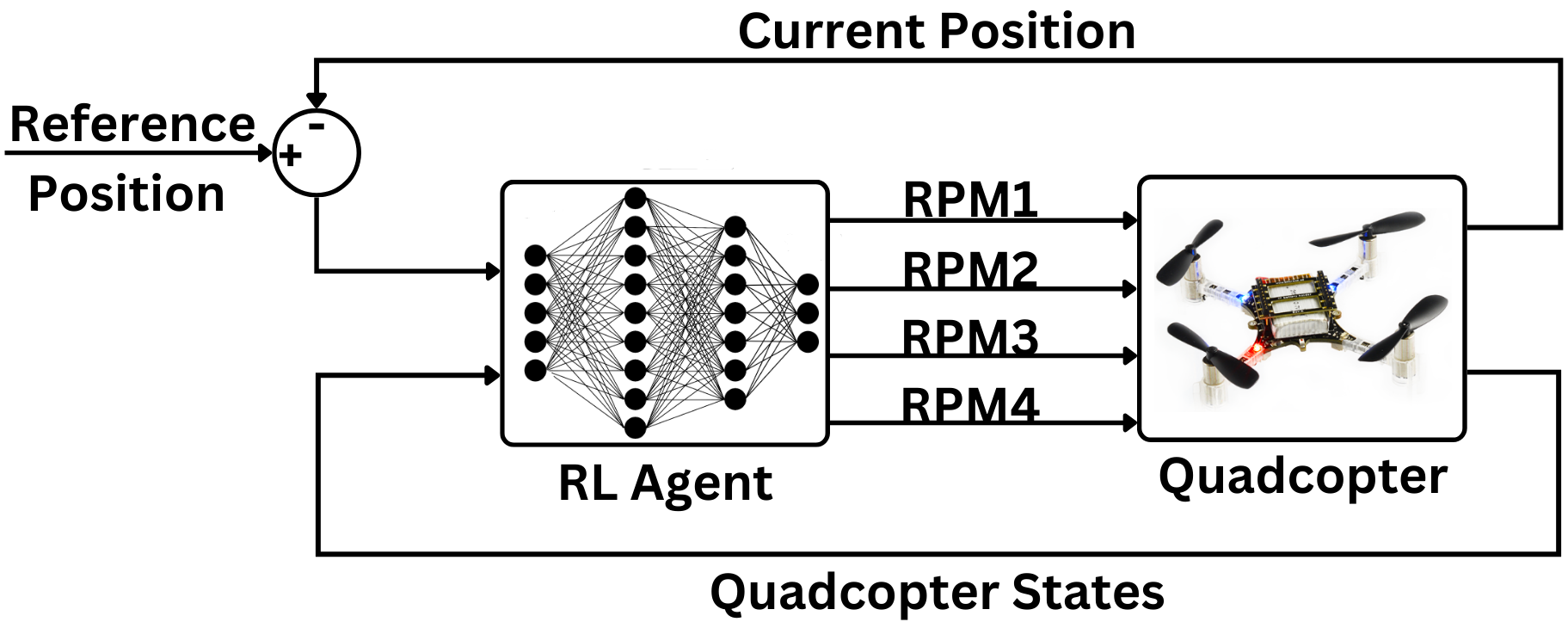}}
\caption{Block diagram of low-level RPM controller}
\label{mokhtar}
\end{figure}

The controller used in this paper is a deep RL agent that controls the motor RPMs directly. The RL agent takes as an input the states of the quadcopter along with the position error and outputs four motor commands to the quadcopter as illustrated in Fig. \ref{mokhtar}.

This control technique has proved its efficiency in the literature before in successfully navigating the quadcopter \cite{mazen} \cite{mokhtar2023autonomous} \cite{HIMANSHU2022281} \cite{DBLP:journals/corr/abs-2010-02293} \cite{DBLP:journals/corr/abs-2109-10488}. Since this framework shows reliable performance, it was used in the purpose of this research. 

The observation space of both algorithms consists of 12 inputs. The current orientation of the quadcopter $(\phi, \theta, \psi)$, current linear velocities $(v_{\textit{x}}, v_{\textit{y}}, v_{\textit{z}})$, current angular velocities $( \omega_{\textit{x}}, \omega_{\textit{y}}, \omega_{\textit{z}})$, and the difference between the quadcopter's position and the target position $(\Delta x, \Delta y, \Delta z)$ as shown in Eq. \ref{obs} respectively.
\vspace{7pt}
    \begin{equation}
        S = [\phi, \theta, \psi, v_{\textit{x}}, v_{\textit{y}}, v_{\textit{z}}, \omega_{\textit{x}}, \omega_{\textit{y}}, \omega_{\textit{z}},\Delta x, \Delta y, \Delta z]
        \label{obs}
    \end{equation}\vspace{7pt}
    
    The action space for both algorithms is continuous. The actions represent the four RPMs of the quadcopter’s rotors as shown in Eq. \ref{action} where ($U_{i}$) is the RPM value. The actions are normalized to the range [-1, 1] and are then mapped to the actual RPM values.

    \begin{equation}
        A = [U_{1}, U_{2}, U_{3}, U_{4}]
        \label{action}
    \end{equation}

\subsection{TD3 and SAC Hyperparameters}
The effectiveness of any RL algorithm is strongly connected to its hyperparameters. A small learning rate was chosen to ensure stability. A high discount factor encourages the agent to favor future rewards over immediate ones. A normal action noise with a standard deviation of 0.2 was added to encourage exploration over exploitation. The same set of hyperparameters is used across the two algorithms. However, the SAC introduces an extra parameter which is the entropy coefficient. To test the effect of dynamic entropy tuning, the entropy coefficient was learned through the environment. The hyperparameters that were used for both agents are shown in Table \ref{tab:ddpg_hyperparameters}.
\vspace{7pt}
    \begin{table}[H]
    \centering
    \caption{Hyperparameters used for the TD3 and SAC agents}
    \begin{tabular}{lll}
    \hline
    Symbol & Description & Value \\
    \hline
    $\lambda$ & Learning rate & $0.0007$ \\
    $B$ & Size of the replay buffer &  $1,000,000$\\
    $-$& Steps before learning starts  & $10,000$ \\

    $N$ & Minibatch size & $256$ \\
    $\tau$ & Update coefficient  & $0.005$ \\
    $\gamma$ & Discount factor & $0.99$ \\
    $-$ & Training frequency  & $1$ episode  \\
    $\sigma_{t}$ & Target policy noise  & $0.2$ \\
    $\sigma_{a}$ & Exploration noise  & $0.2$ \\
    $-$ & Target noise clip & $0.5$ \\
    $f$ & Agent frequency  & $50$ Hz \\
    $t_{max}$  & Maximum steps of one episode  & $502$ \\
    \hline
    \end{tabular}
    \label{tab:ddpg_hyperparameters}
    \end{table}
    \vspace{7pt}
    
\subsection{Network Training}
The training procedure followed the conditions and parameters demonstrated above. The training platform was a CrazeFlie2.0 quadcopter \cite{8046794} model set up in an X layout in the gym-pybullet-drones environment \cite{pybullet}. The PyTorch library, which smoothly incorporates CUDA, was used to construct RL algorithms, while the Stable-Baselines3 library made it easier to implement the TD3 and SAC algorithms. TensorBoard was used to track progress. Training was accelerated by using NVIDIA CUDA on a GeForce RTX 3070 GPU, which allowed for the quick completion of multiple episodes.
\vspace{7pt}

\section{Results}
In this section, the results of stochastic algorithms and deterministic algorithms are compared. To fully test the deterministic and stochastic algorithms and their exploration efficiency, the agents were trained using two different boundary conditions.

\subsection{Small Environment Training}
The objective of this training was for the agent to stabilize the quadcopter at a fixed point of [0,0,1] starting from a random position in each episode. Both the stochastic and deterministic training ran for a total of 4 million steps. A small environment is used to test the exploration efficiency over a small bound. The small bounds were also used to test whether the algorithms will be able to generalize the policy over a small environment. The initial position was chosen randomly from the following limits:
   \begin{itemize}
        \item \(x\) and \(y\) \(\in\) [-0.5, 0.5]
        \item \(z\) \(\in\) [0.5, 1.5]
    \end{itemize}

\begin{figure}[H]
\centerline{\includegraphics[width=0.47\textwidth]{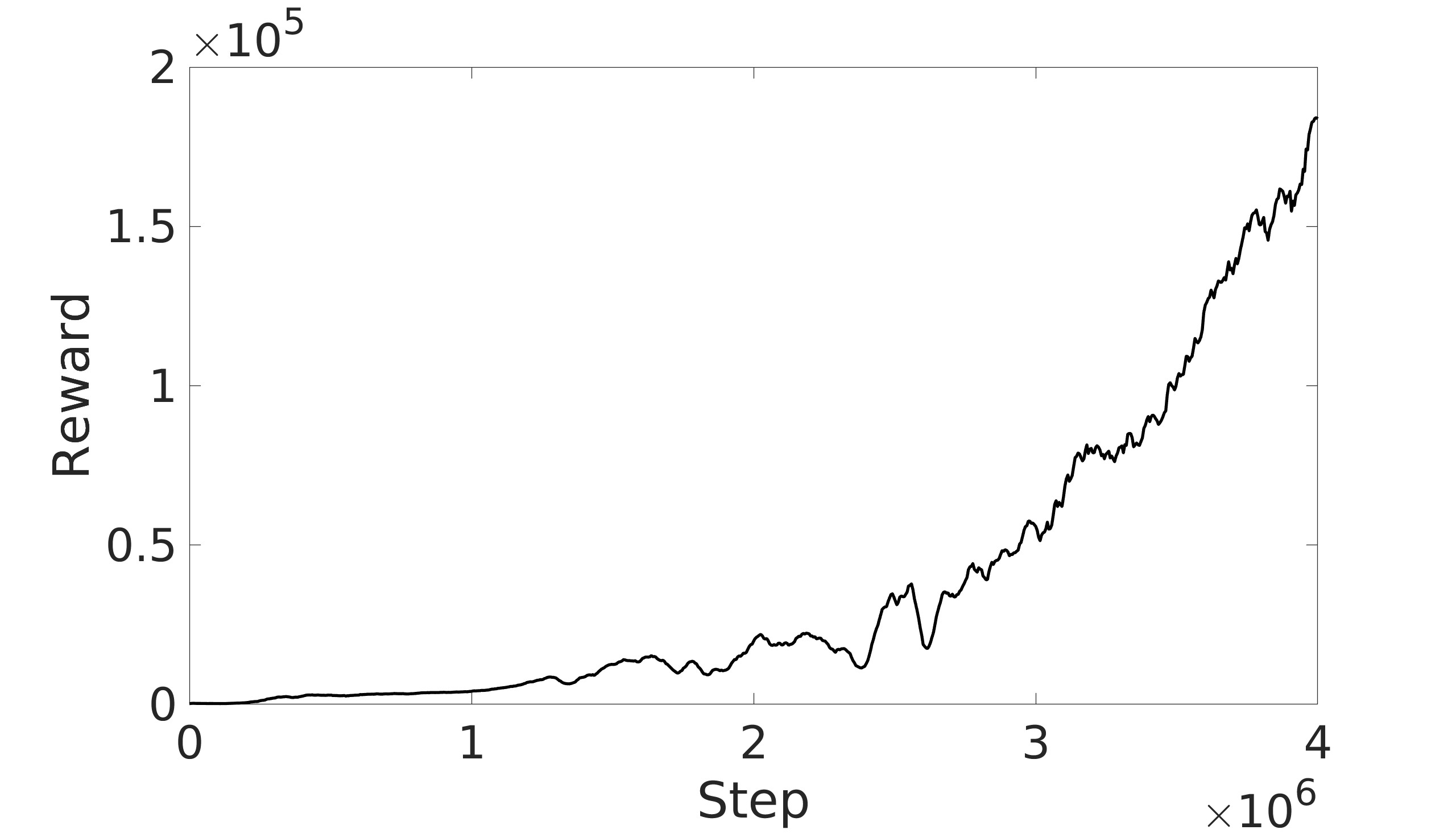}}
\caption{Average mean reward of small environment training for deterministic agent}
\label{td3rew}
\end{figure}

Fig. \ref{td3rew} shows the mean reward obtained by the deterministic agent. The deterministic agent reached a maximum reward of around 175,000 at the end of the training. An increasing learning curve is shown with small variations in performance as the curve is not smooth.

\begin{figure}[H]
\centerline{\includegraphics[width=0.47\textwidth]{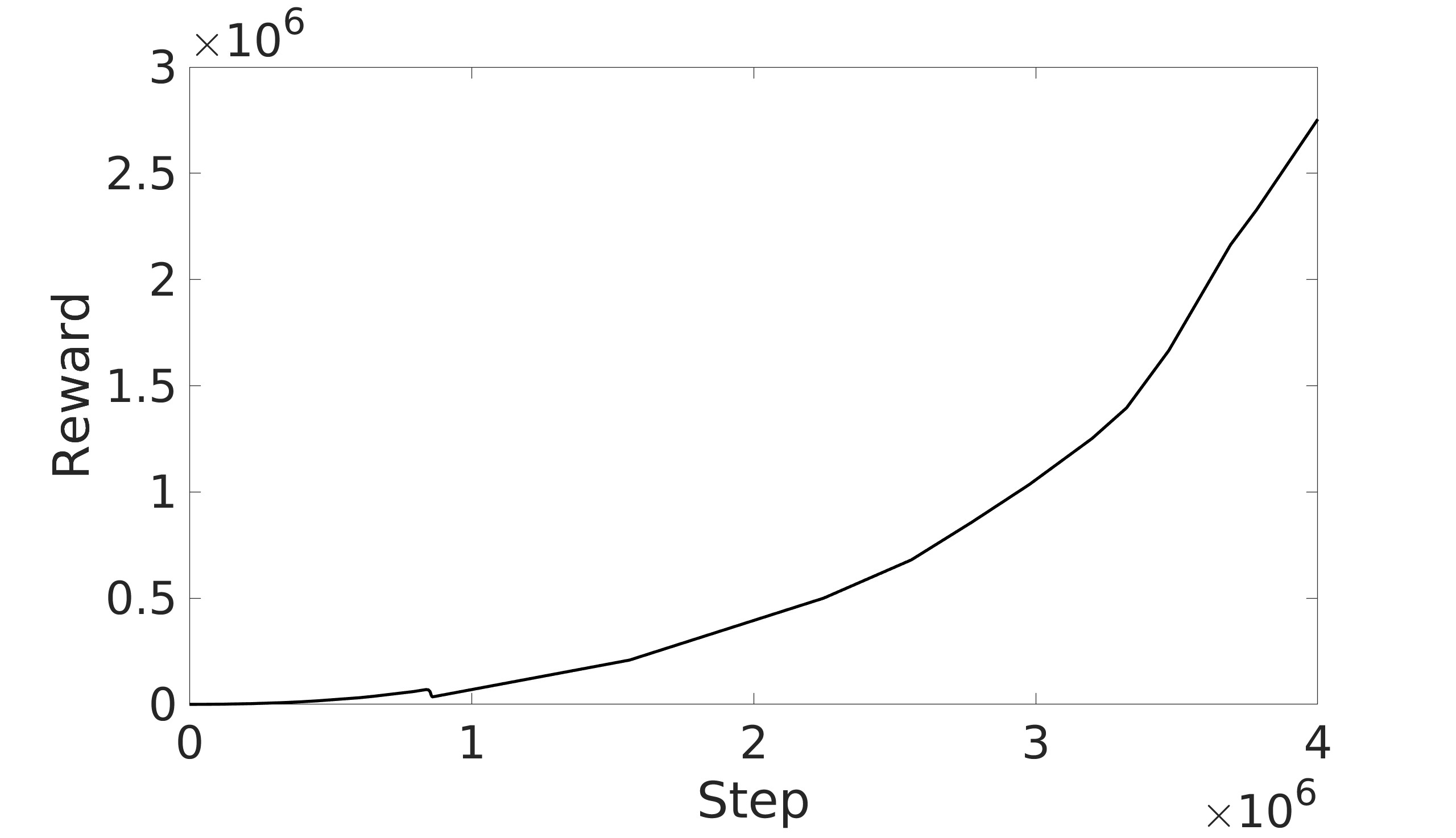}}
\caption{Average mean reward of small environment training for stochastic agent}
\label{sacrew}
\end{figure}

Fig. \ref{sacrew} shows the mean reward obtained by the stochastic agent. The stochastic agent reached a higher maximum reward than the deterministic agent. The maximum reward is around 2,750,000 at the end of the training. This is around 10 times higher than the deterministic agent. A smoother learning curve is shown with almost no variations in performance.

\begin{figure}[H]
\centerline{\includegraphics[width=0.47\textwidth]{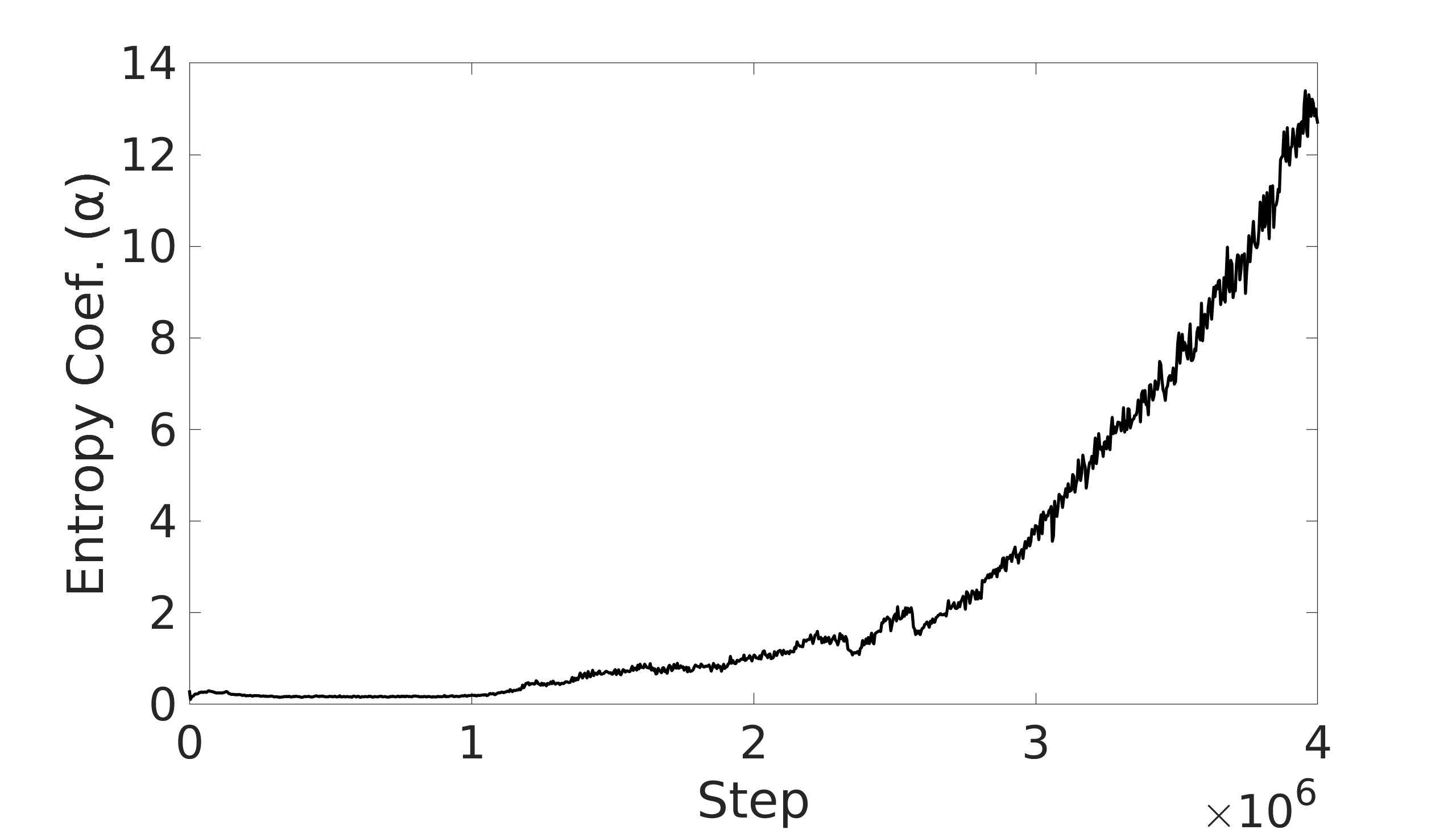}}
\caption{Entropy coefficient ($\alpha$) values of small environment training for stochastic agent}
\label{entsm}
\end{figure}

Fig. \ref{entsm} shows the entropy coefficient ($\alpha$) values at each time step for the stochastic agent. The entropy curve takes the same shape of the reward curve. The entropy values are increasing with time along with the reward showing that the maximum entropy RL goal is achieved.

\begin{figure}[H]
            \centerline{\includegraphics[width=0.47\textwidth]{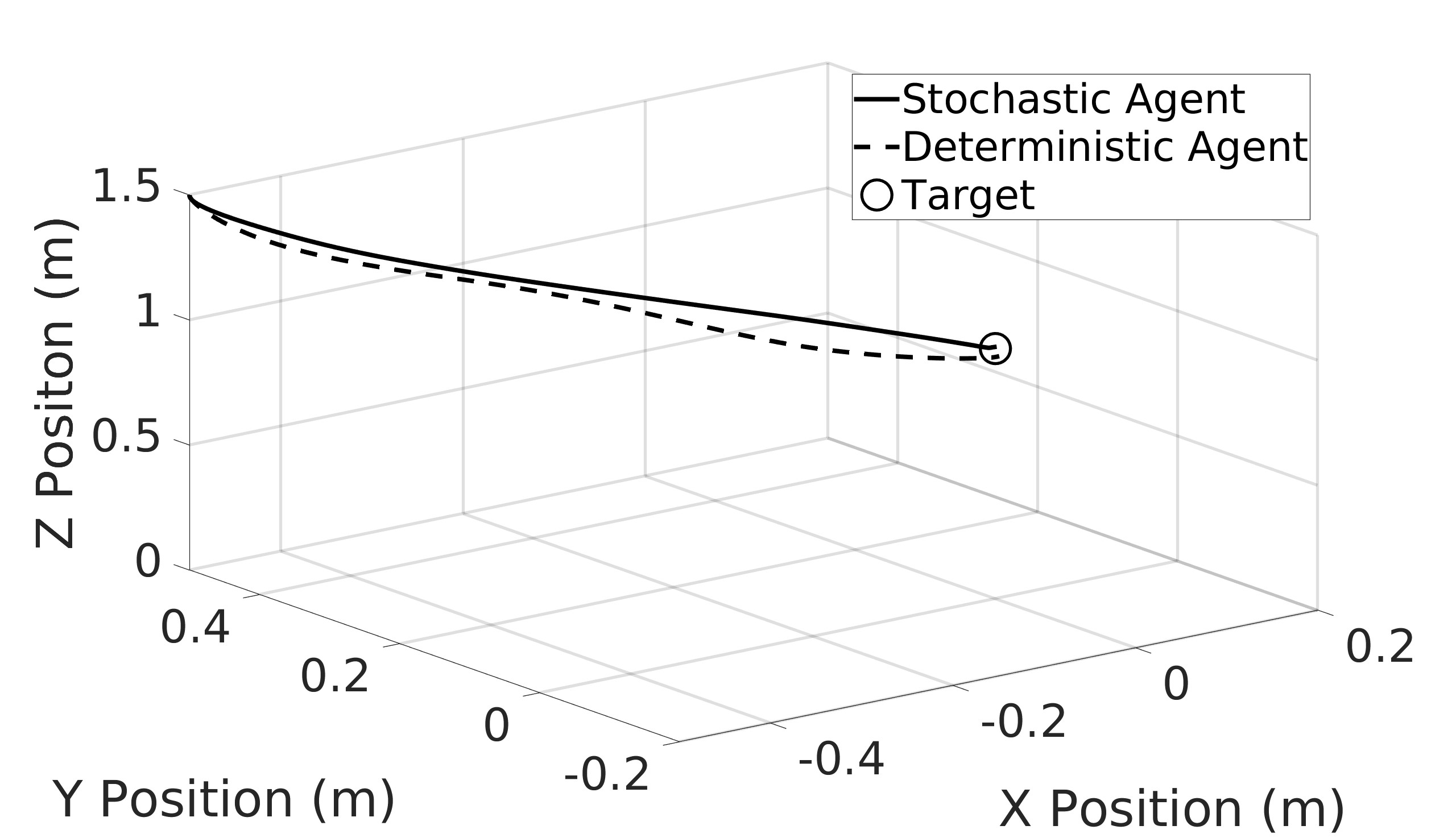}}
            \caption{Response of deterministic and stochastic agents in small environment from an initial position of [-0.5, 0.5, 1.5]}
            \label{x0.5}
    \end{figure}

Fig. \ref{x0.5} shows the stochastic agent response when starting from an initial position of [-0.5, 0.5, 1.5] compared to the deterministic agent response. This point was chosen as the most extreme position the agents were trained on. Both agents stabilized with zero steady-state errors.
    
\begin{figure}[H]
            \centering
            \includegraphics[width=0.47\textwidth]{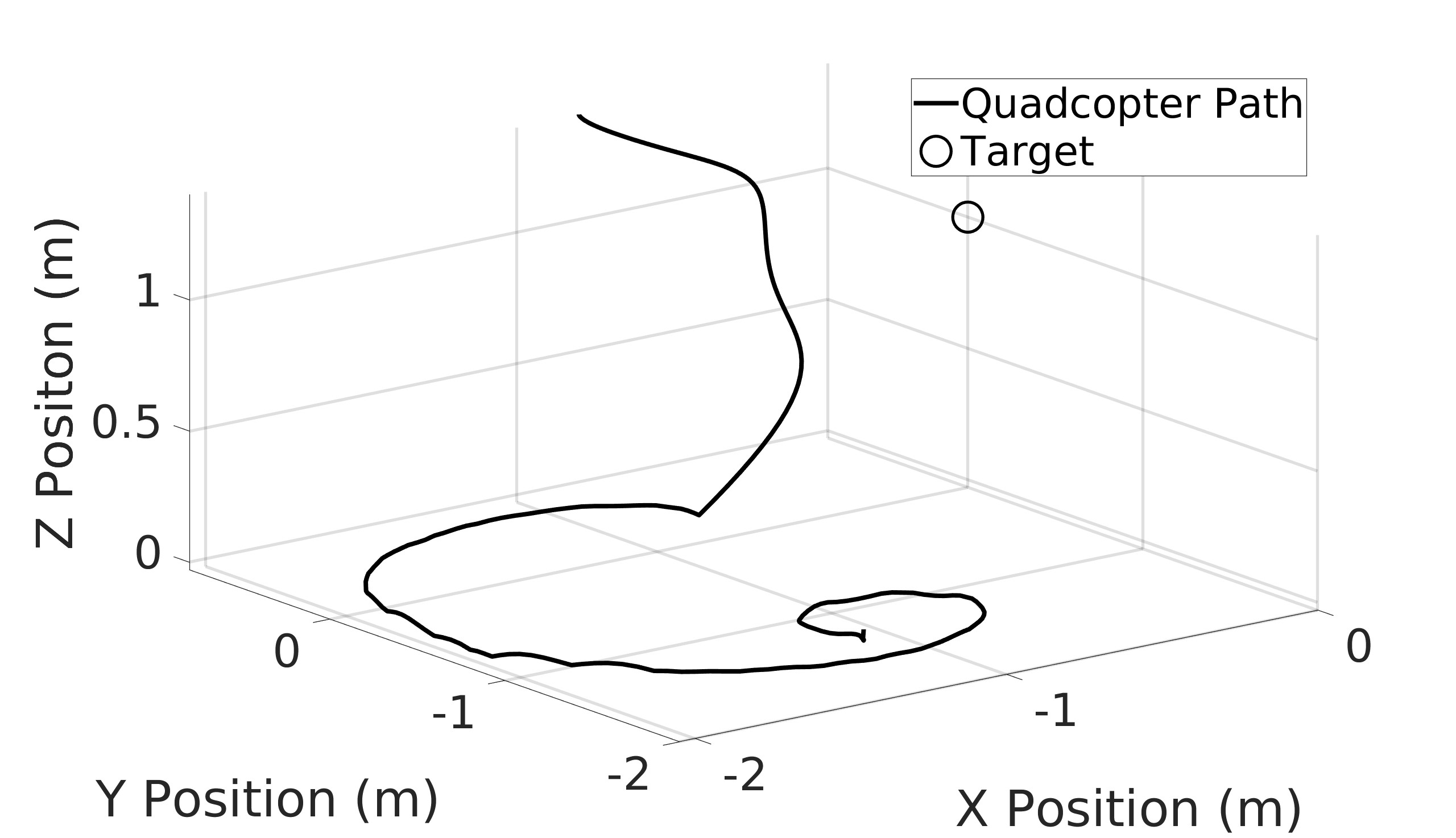}
            \caption{Response of deterministic agents in small environment from an initial position of [-0.8, 0.8, 1.4]}
            \label{offT0.5}
    \end{figure}

        The real difference between the two algorithms appears when passing initial positions to the agent outside the trained environment. The deterministic agent crashed almost instantly as shown in Fig. \ref{offT0.5}.
    
\begin{figure}[H]
            \centering
            \includegraphics[width=0.5\textwidth]{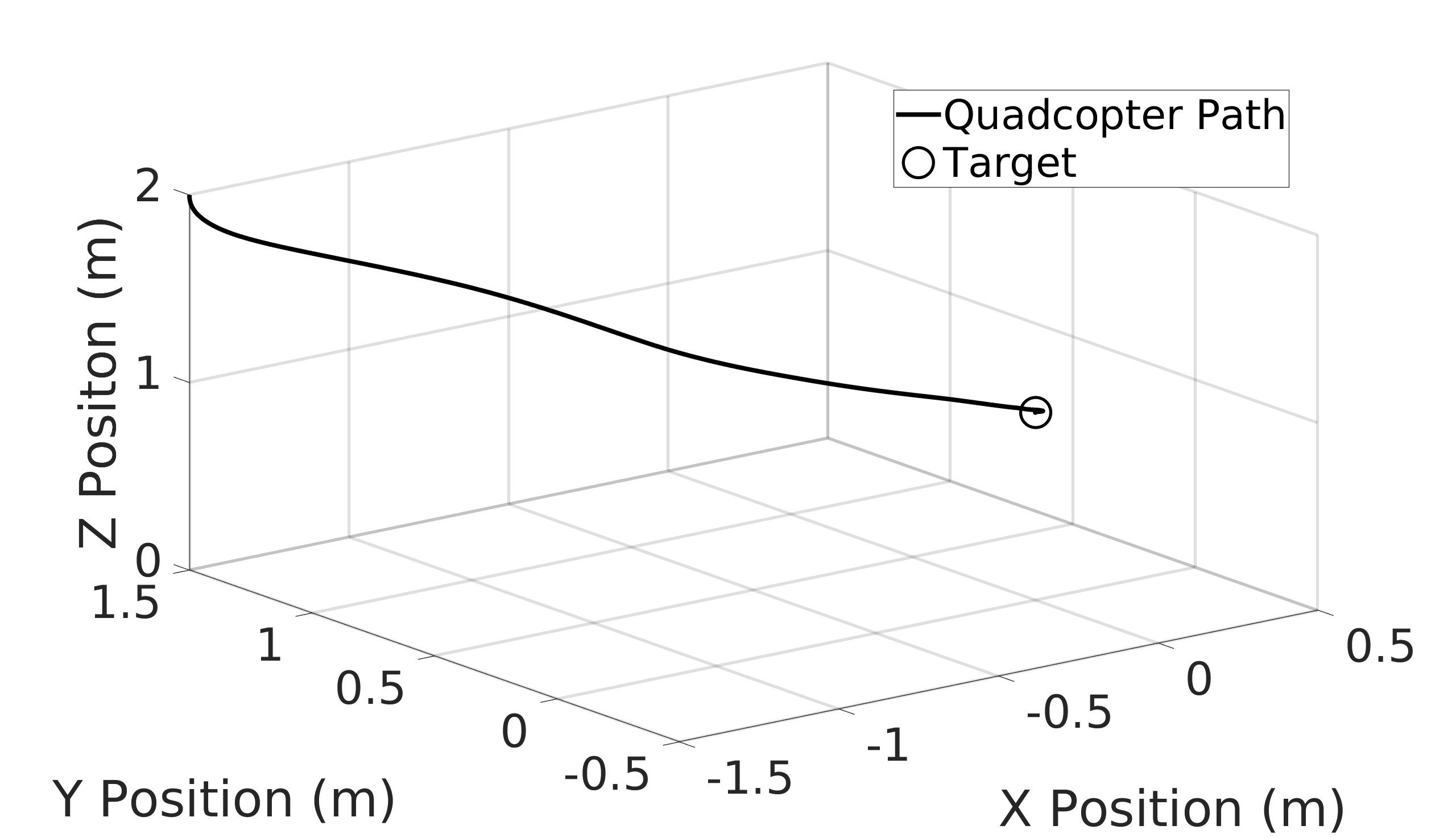}
            \caption{Response of stochastic agent in small environment from an initial position of [-1.5, 1.5, 2]}
            \label{offS0.5}
    \end{figure}

    Fig. \ref{offS0.5} shows the response of the stochastic agent when starting with an initial position of [-1.5, 1.5, 2]. The stochastic agent successfully stabilized at the target position as shown although the agent was never trained on this point at all. This shows the stochastic agent's ability to generalize the policy and achieve better performance than the deterministic algorithm.
    
 \subsection{Large Environment Training}
In this training, the agents were trained with a larger environment with the same goal. The larger environment was used to test the exploration efficiency of both algorithms in exploring and learning a larger environment. The algorithms trained for the same time frame of four million steps. The initial position limits were increased to the following values:
    \begin{itemize}
        \item \(x\) and \(y\) \(\in\) [-2.5, 2.5]
        \item \(z\) \(\in\) [0.2, 2.5]
    \end{itemize}
    
The deterministic algorithms train a sample-efficient policy where each state has a single deterministic action, exploration is done off-policy by adding external noise to the actions. To test this exploration technique, two different Gaussian noises were added to two different agents as shown below. 
\begin{itemize}
    \item Deterministic Agent with relatively low Gaussian Noise with a mean of 0.2 $\big(N(0,0.2)\big)$
    \item Deterministic Agent with relatively high Gaussian Noise with a mean of 0.5 $\big(N(0,0.5)\big)$
\end{itemize}

Each agent was trained seperatly and training results are shown below. Fig. \ref{low} shows the average mean reward obtained by the deterministic agent with relatively low noise. The agent reached an optimal policy then catastrophic forgetting started to happen. The performance of the agent gradually worsened and high variations in performance between each episode started to appear.
\begin{figure}[H]
            \centering
            \includegraphics[width=0.5\textwidth]{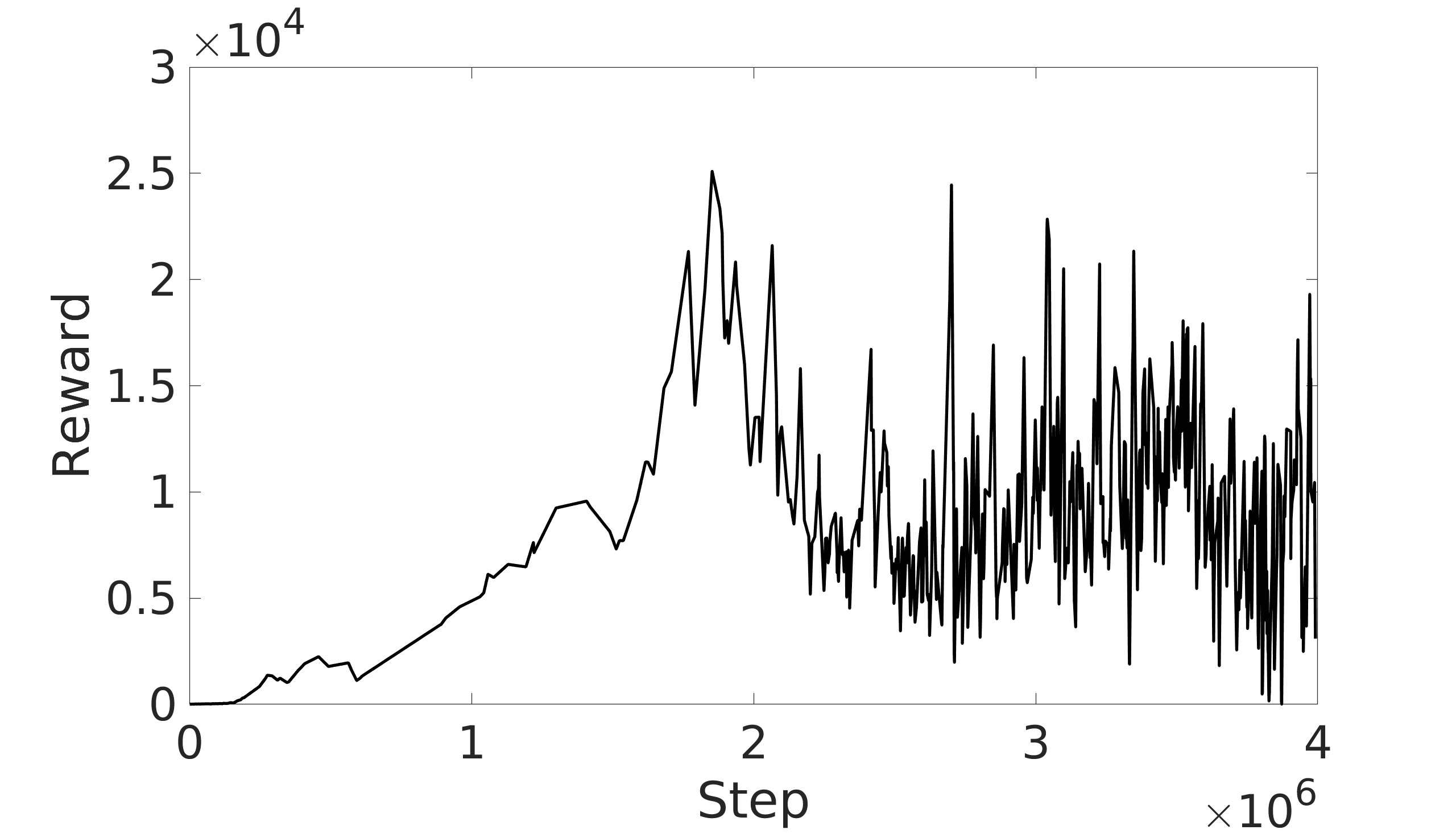}
            \caption{Average mean reward of large environment training for deterministic agent with relatively low noise}
            \label{low}
            \end{figure}
            
\begin{figure}[H]
            \centering
            \includegraphics[width=0.5\textwidth]{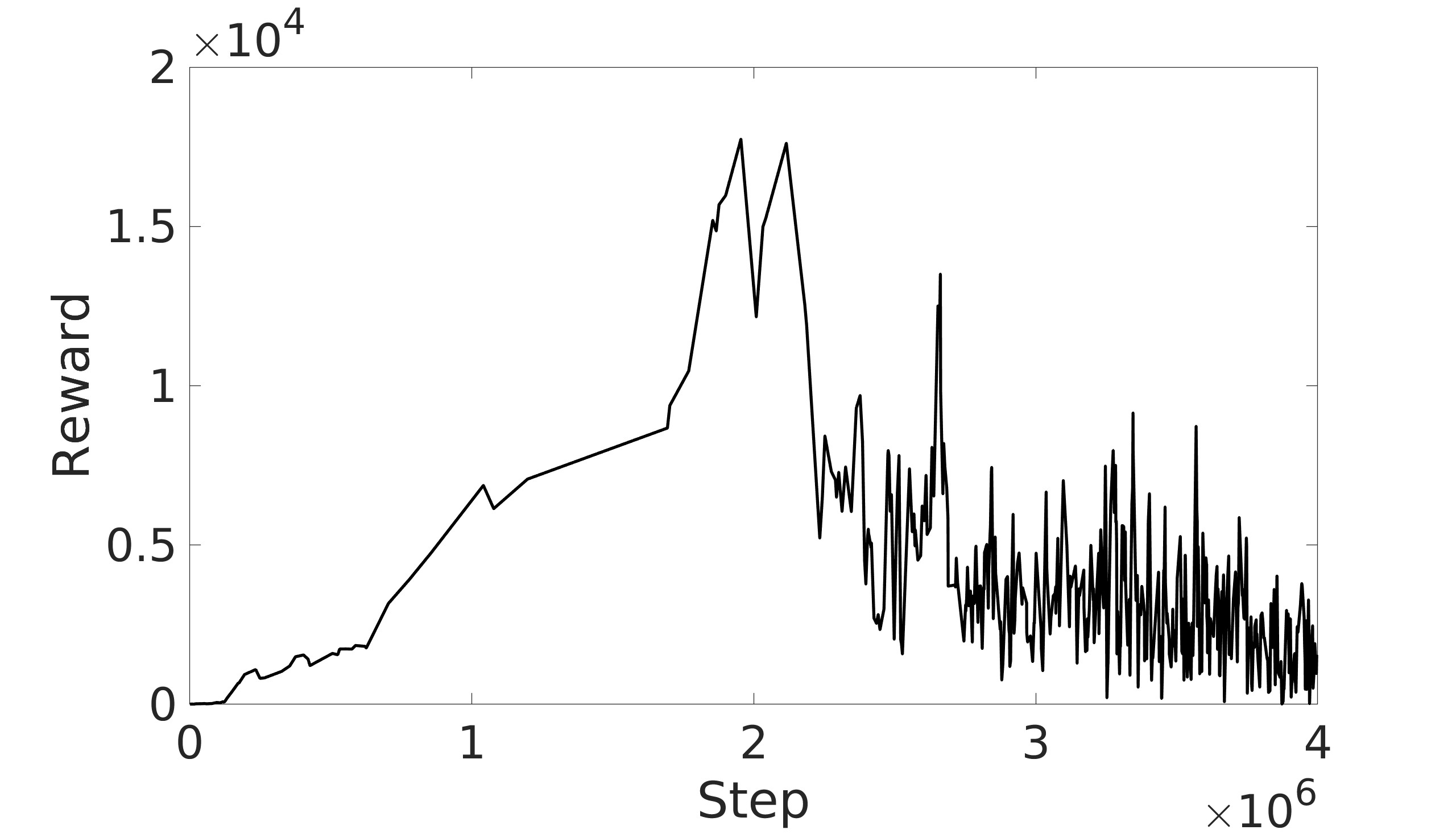}
            \caption{Average mean reward of large environment training for deterministic training with relatively high noise}
            \label{high}
    \end{figure}

Deterministic algorithms gradually worsen after a peak performance as learning continues, this might occur since with a better policy, experiences that have never occurred in the system before are discovered \cite{lu2020adaptive}. Another deterministic agent was trained using a higher normal action noise with a mean of 0.5 to test if higher noise would result in better exploration. However, as shown in Fig. \ref{high}, the catastrophic forgetting worsened as the noise increased and higher variations in performance between each episode were present.

    The external noise approach is wasteful and leads to unstable behavior since some regions in the action space close to the current policy are likely to have already been explored by past policies leading to sub-optimal policies \cite{NEURIPS2019_a34bacf8}.

    To test the effect of dynamic entropy tuning on exploration efficiency, three different scenarios were explored as shown below. 
    \begin{itemize}
    \item Stochastic Agent with Static Entropy
    \item Stochastic Agent with Dynamic Entropy and added Gaussian Noise with a mean of 0.2 $\big(N(0,0.2)\big)$
    \item Stochastic Agent with Dynamic Entropy Tuning Only
\end{itemize}

\begin{figure}[H]
            \centering
            \includegraphics[width=0.5\textwidth]{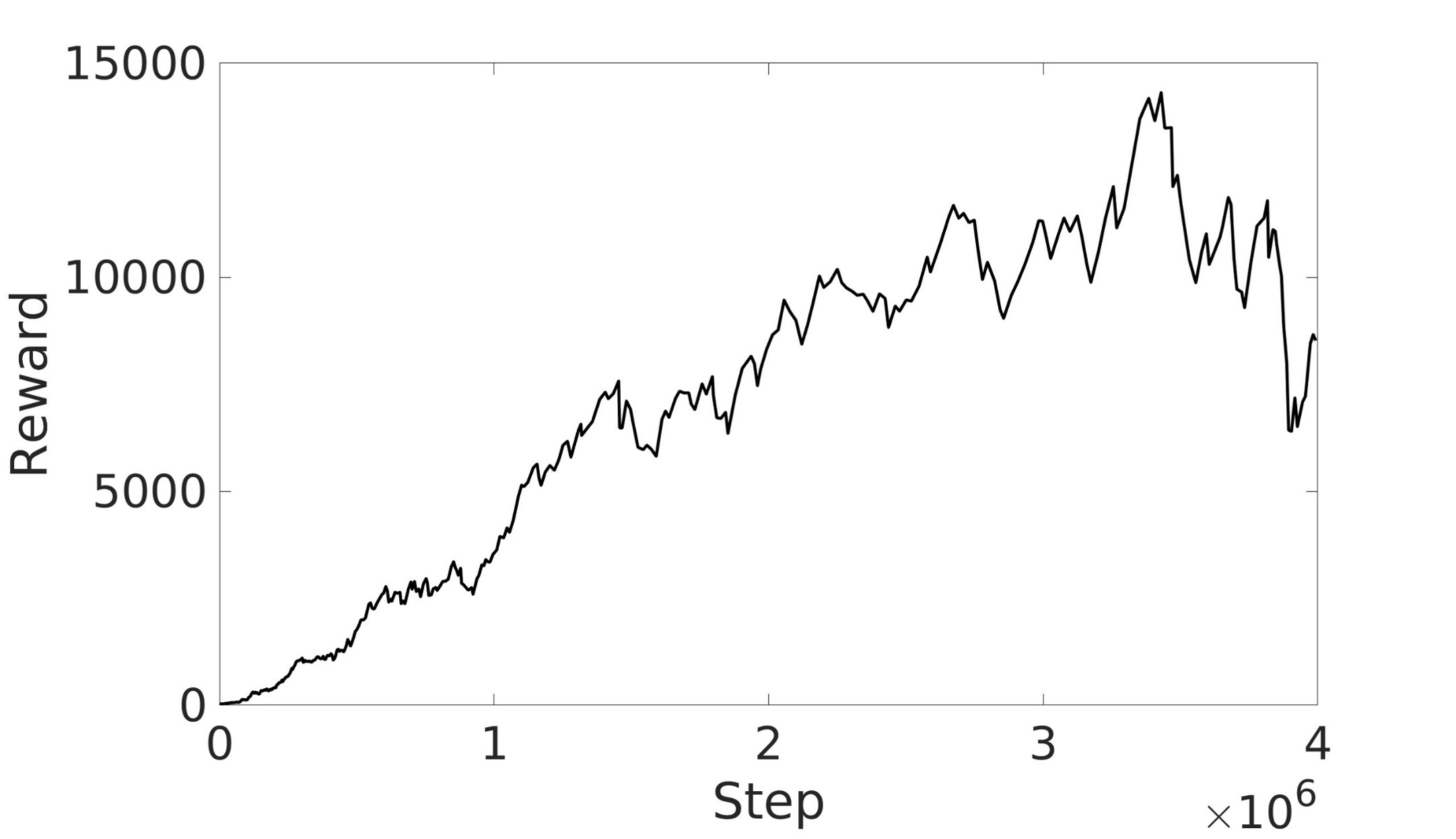}
            \caption{Average mean reward of large environment training for stochastic agent with static entropy}
            \label{highn}
    \end{figure}

    In the first scenario, a constant entropy coefficient was used to see whether dynamic entropy tuning has a better impact than static entropy. Fig. \ref{highn} shows the mean reward obtained by the agent. Compared to the other scenarios, the maximum reward is relatively low and the constant entropy failed to aid in exploration. This proves that choosing an optimal entropy value is a nontrivial task \cite{haarnoja2019soft}.
    \vspace{7pt}
    \vspace{7pt}
    
\begin{figure}[H]
            \centering
            \includegraphics[width=0.5\textwidth]{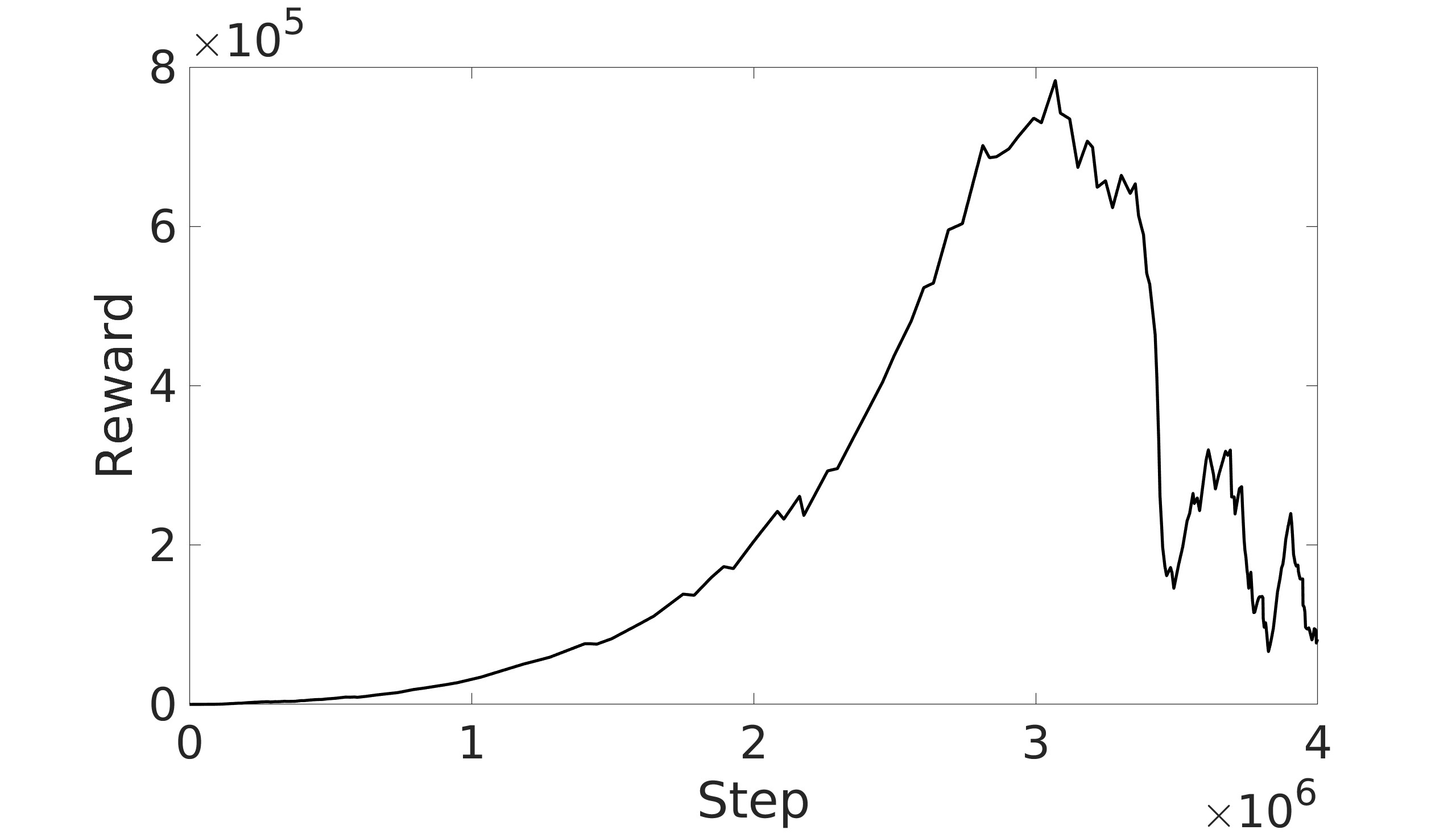}
            \caption{Average mean reward of large environment training for stochastic agent with dynamic entropy and added noise}
            \label{noise}
    \end{figure}

    Dynamic entropy tuning with added external noise was explored next to see if this pairing would result in better exploration. This pairing was also used to test whether the catastrophic forgetting in deterministic algorithms is caused by the noise or a flaw in the algorithm itself. Fig. \ref{noise} shows the mean reward obtained by the stochastic training agent. The agent reached a higher reward than all previous results. However, catastrophic forgetting started to appear after reaching the optimal policy. This proves that adding external noise to the output leads to a fast catastrophic forgetting after reaching an optimal policy.
    \vspace{7pt}
    
    \vspace{7pt}
    
\begin{figure}[H]
            \centering
            \includegraphics[width=0.5\textwidth]{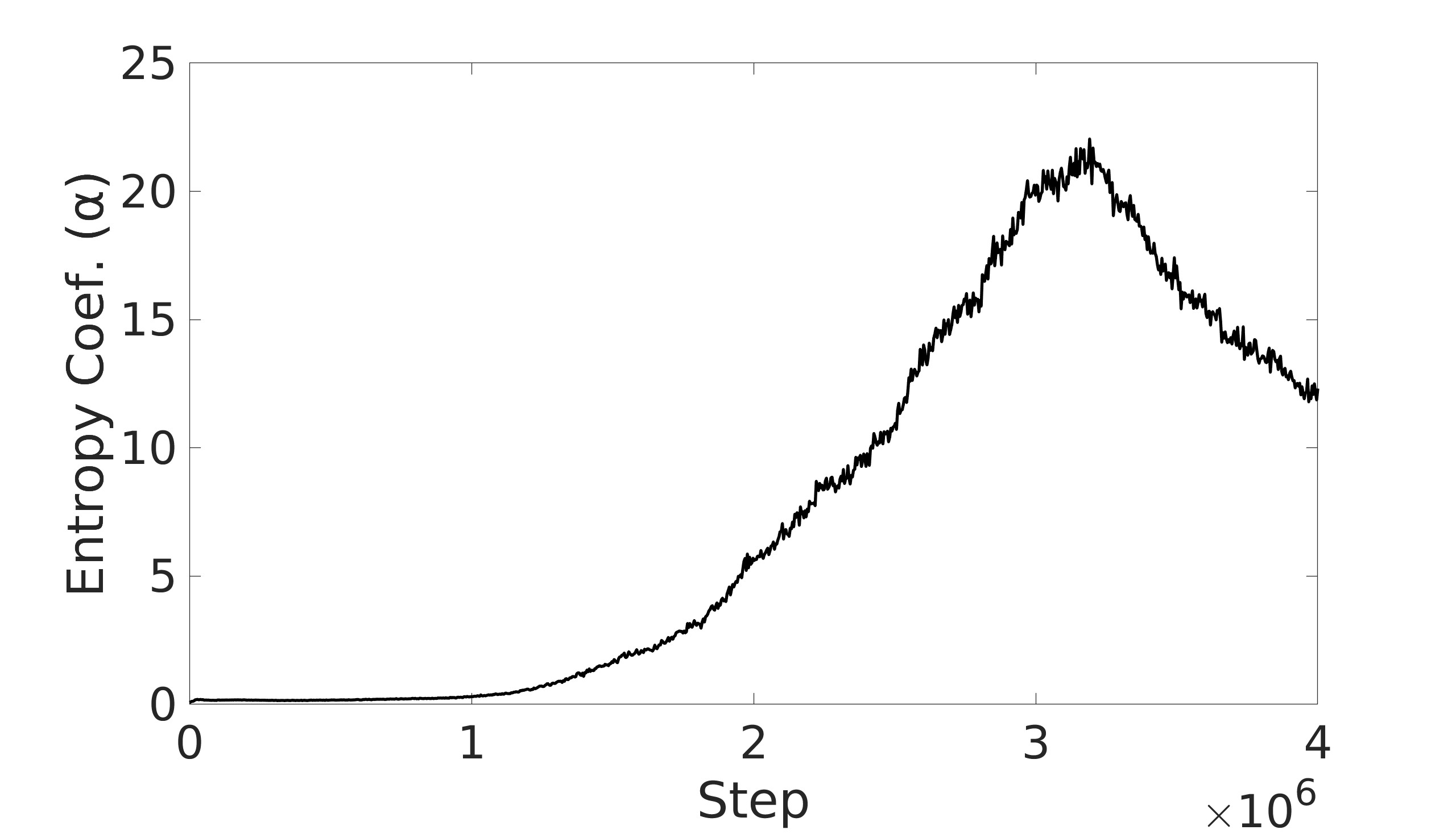}
            \caption{Entropy coefficient ($\alpha$) values of large environment training for stochastic training with dynamic entropy and added noise}
            \label{entn}
    \end{figure}

\begin{figure}[H]
            \centering
            \includegraphics[width=0.4\textwidth]{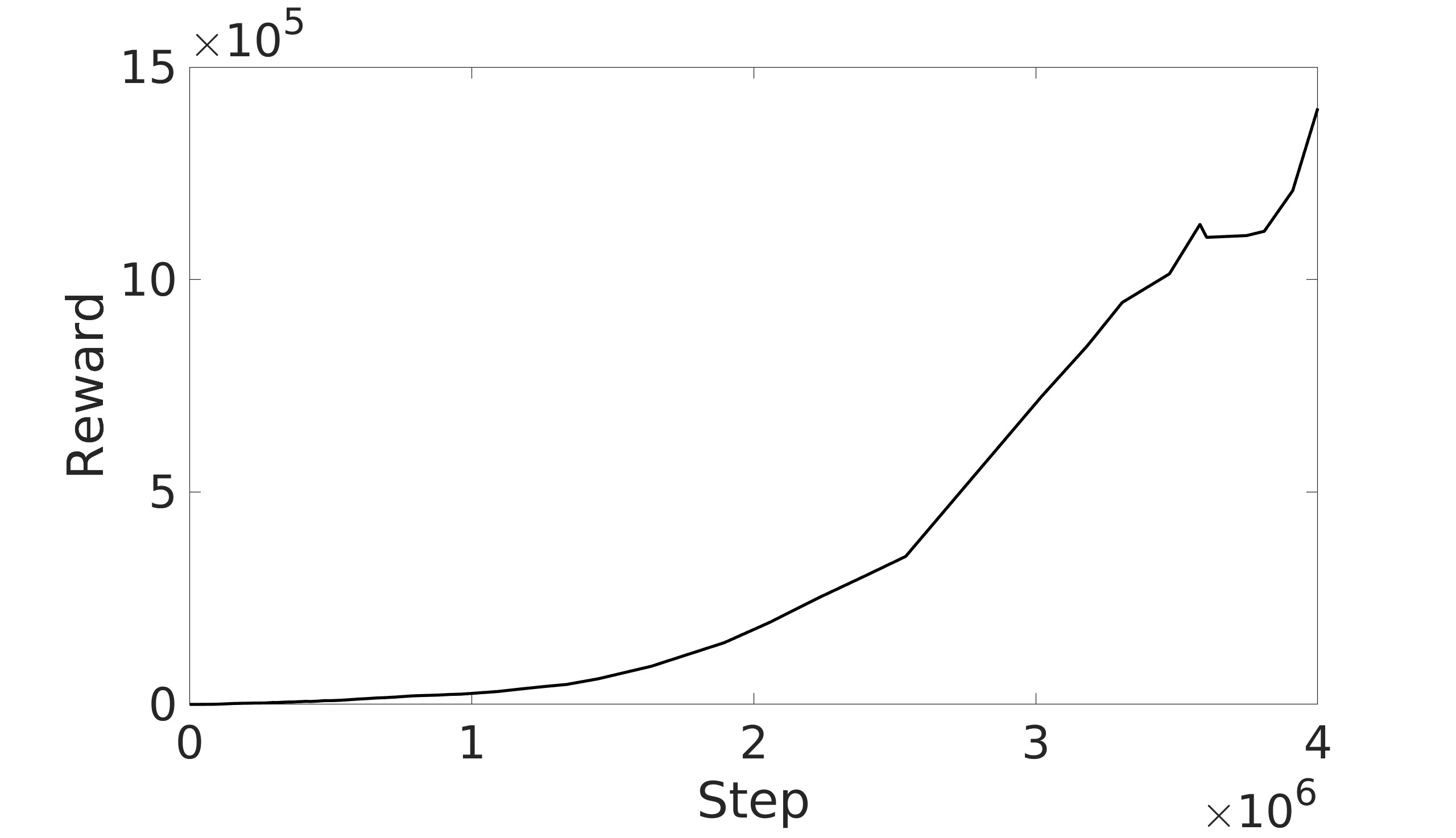}
            \caption{Average mean reward of large environment training for stochastic agent with dynamic entropy only}
            \label{nnoise}
    \end{figure}

    Fig. \ref{entn} shows the entropy coefficient ($\alpha$) values of the stochastic agent with added external noise. The entropy values started to decrease with the catastrophic forgetting. Both the rewards and entropy values decreased with time showing that both goals of maximum entropy RL, maximizing both rewards and entropy, were not achieved.
    \vspace{7pt}
    \vspace{7pt}
    Finally, dynamic entropy tuning without added external noise was tested. Fig. \ref{nnoise} shows the mean reward obtained by the stochastic training agent. The agent was trained for the same number of steps and reached a higher reward than all previous trainings. A smoother reward curve is achieved and the optimal policy was reached without catastrophic forgetting happening. This result show the better exploration efficiency the dynamic entropy tuning allows the stochastic algorithms to have. It also helps in a more stable training procedure.
    \vspace{7pt}
    \vspace{7pt}
    The absence of catastrophic forgetting from the reward curve further proves that adding external noise to output is inefficient and a better exploration benefit can be reached using only dynamic entropy tuning while also avoiding the catastrophic forgetting that happens due to the external noise.
    \vspace{7pt}
    
\begin{figure}[H]
            \centering
            \includegraphics[width=0.4\textwidth]{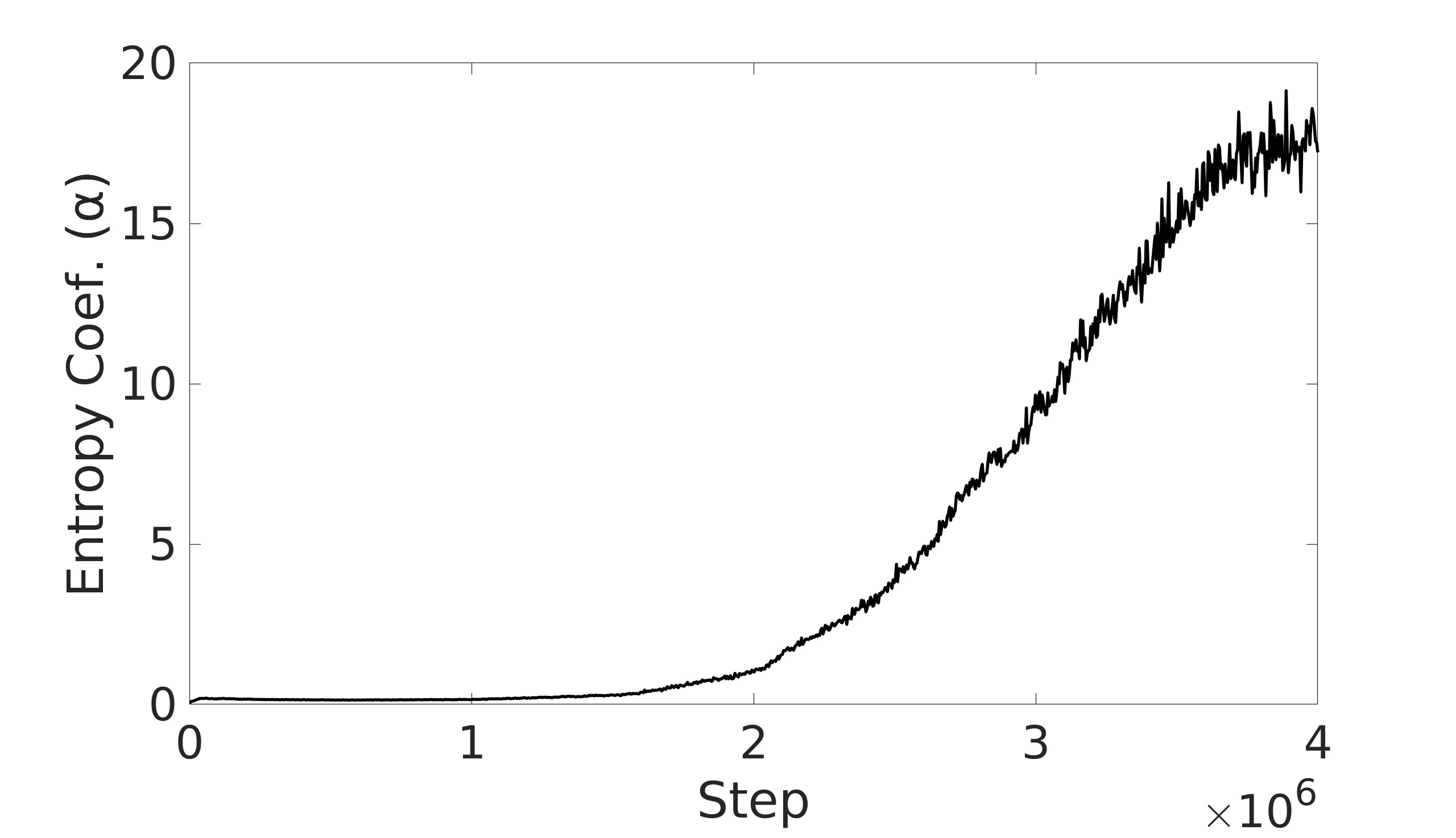}
            \caption{Entropy coefficient ($\alpha$) values of large environment training for stochastic agent with dynamic entropy only}
            \label{entnn}
    \end{figure}

    Fig. \ref{entnn} shows the entropy coefficient ($\alpha$) values of the stochastic agent with dynamic entropy tuning. The entropy coefficient ($\alpha$) values were increasing without any decrease in values. The increase in both the rewards and entropy coefficient ($\alpha$) values shows that the goal of maximum entropy RL is achieved with the dynamic entropy tuning.

\begin{figure}[H]
            \centering
            \includegraphics[width=0.5\textwidth]{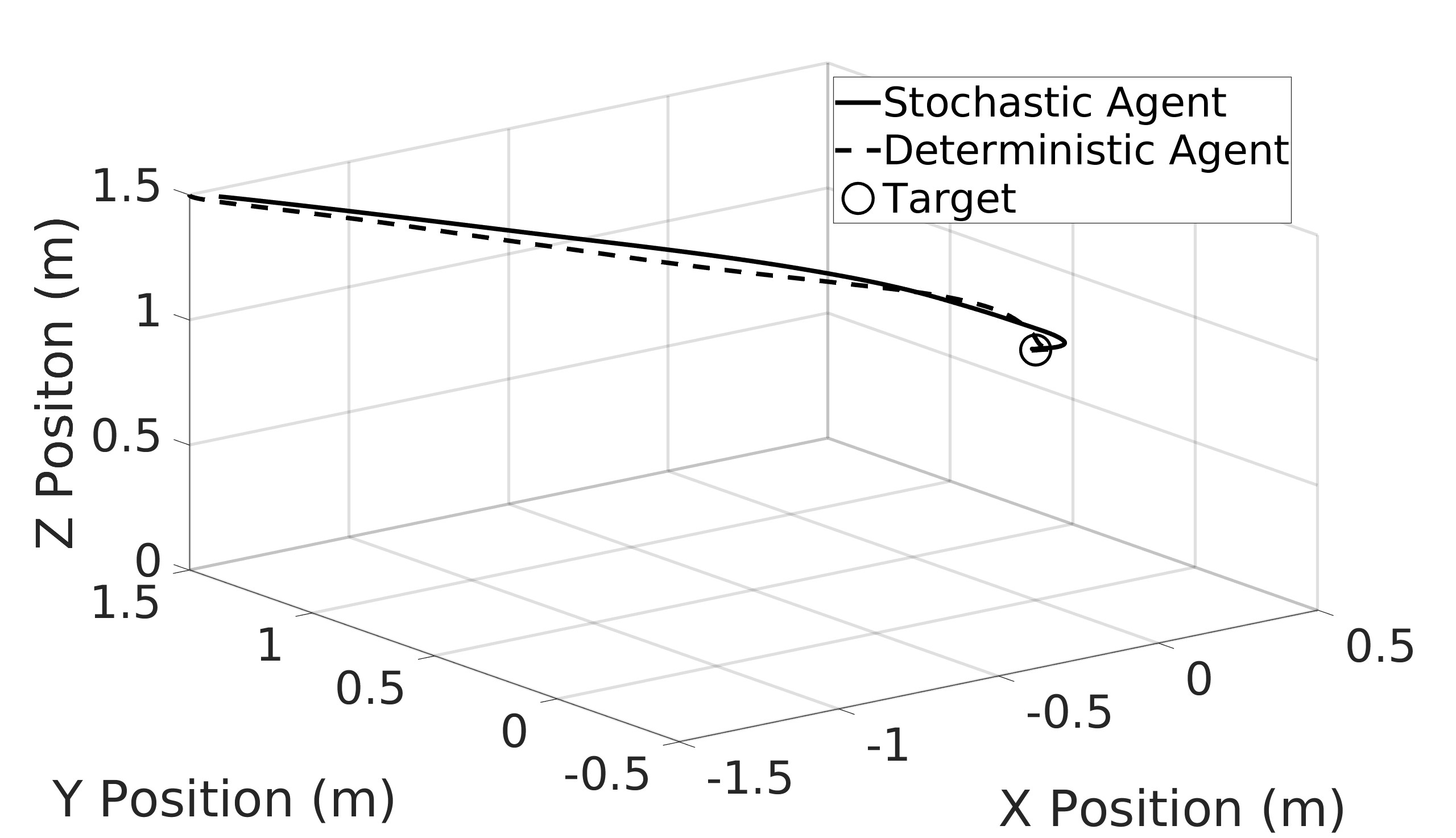}
            \caption{Response of deterministic and stochastic agents in large environment from an initial position of [-1.5, 1.5, 1.5]}
            \label{x2.5}
    \end{figure}

\begin{figure}[H]
            \centering
            \includegraphics[width=0.5\textwidth]{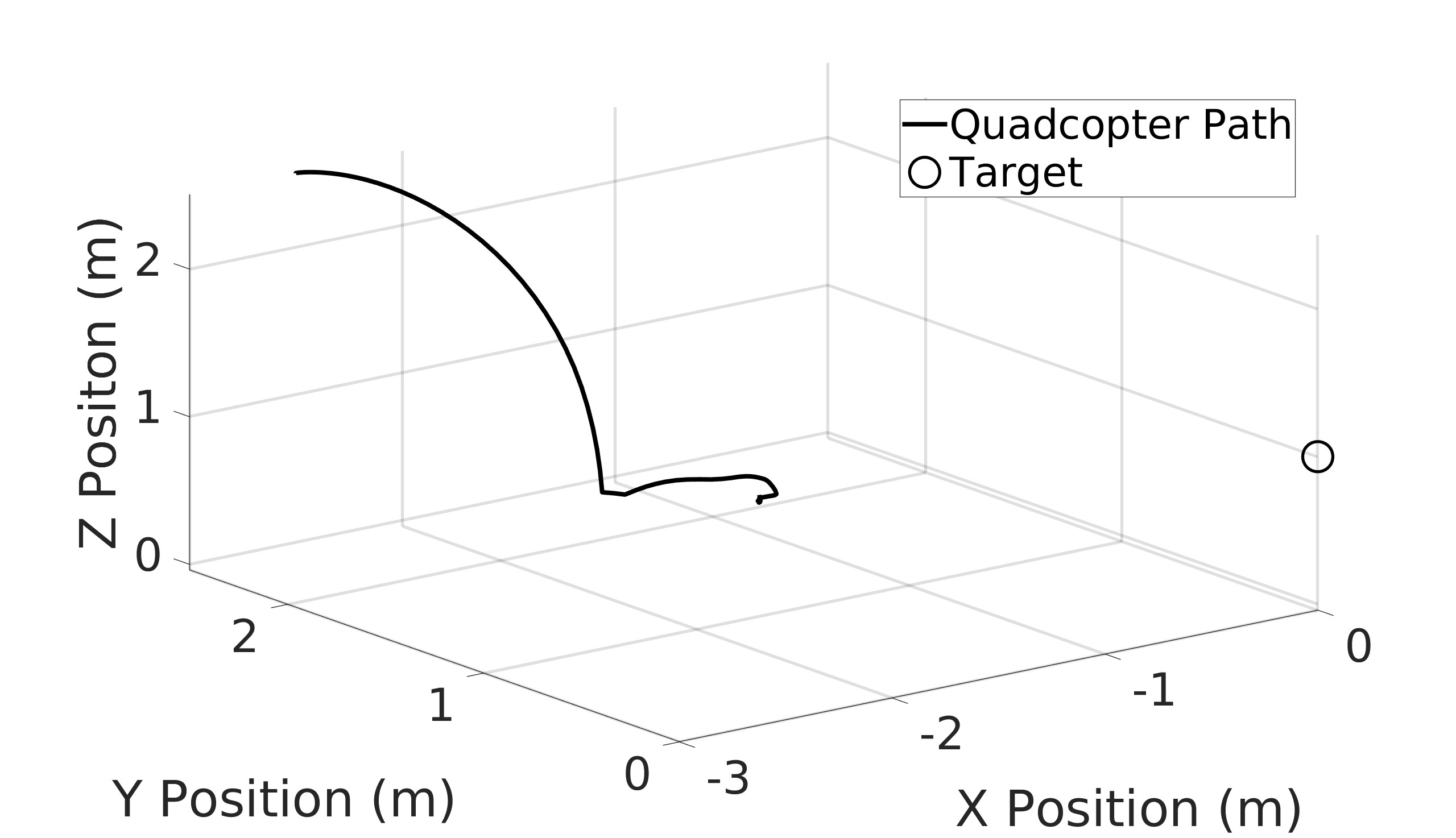}
            \caption{Response of deterministic agent in large environment from an initial position of [-2.5, 2.5, 2.5]}
            \label{T2.5}
    \end{figure}

\begin{figure}[H]
            \centering
            \includegraphics[width=0.5\textwidth]{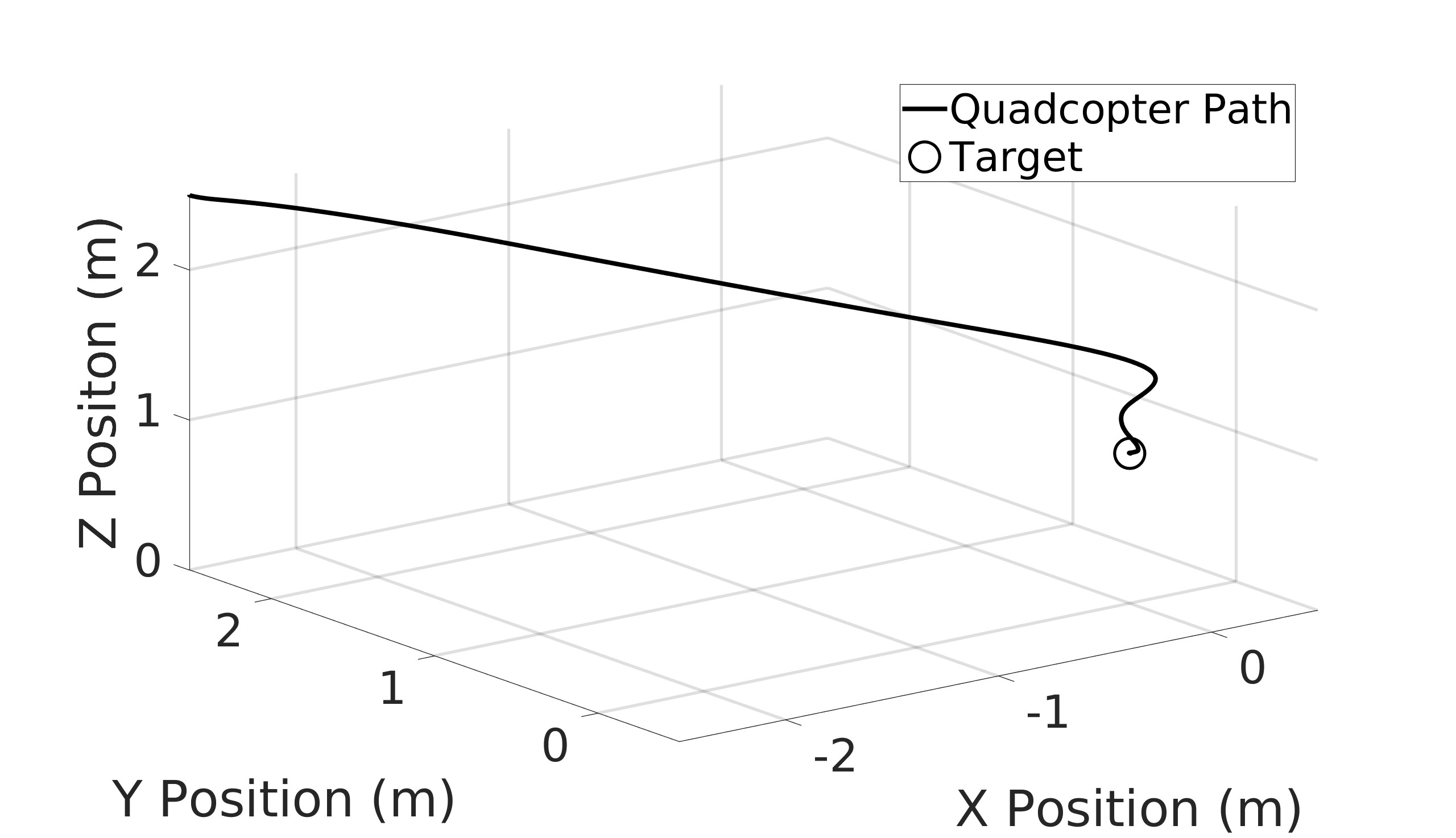}
            \caption{Response of stochastic agent in large environment from an initial position of [-2.5, 2.5, 2.5]}
            \label{S2.5}
    \end{figure}

Fig. \ref{x2.5} shows the response of the best stochastic agent that used dynamic entropy tuning compared to the response of the best deterministic agent that used external noise with a mean value of 0.2. Both starting from an initial position of [-1.5, 1.5, 1.5]. This point was chosen as a close position to the target. Both agents stabilized with zero steady-state errors. As shown the performance of both agents again is almost identical with negligible differences between the two algorithms.
    However, when passing the most extreme point both agents were trained on, point [-2.5, 2.5, 2.5], the deterministic agent crashes and fails to reach the target as shown in Fig. \ref{T2.5}. This shows that the deterministic agent failed to learn the entirety of the environment and had a poor exploration efficiency. 
    Fig. \ref{S2.5} shows the response of the stochastic training agent when starting with an initial position of [-2.5, 2.5, 2.5]. The stochastic training agent successfully stabilized at the target position as shown. 
     The stochastic algorithm was able to learn the entire environment faster than the deterministic algorithm due to the dynamic entropy tuning giving the stochastic algorithm more efficient exploration techniques.
     Catastrophic forgetting also prevented the deterministic algorithm from continuing learning in the environment and halted the learning process.

\begin{figure}[H]
            \centering
            \includegraphics[width=0.47\textwidth]{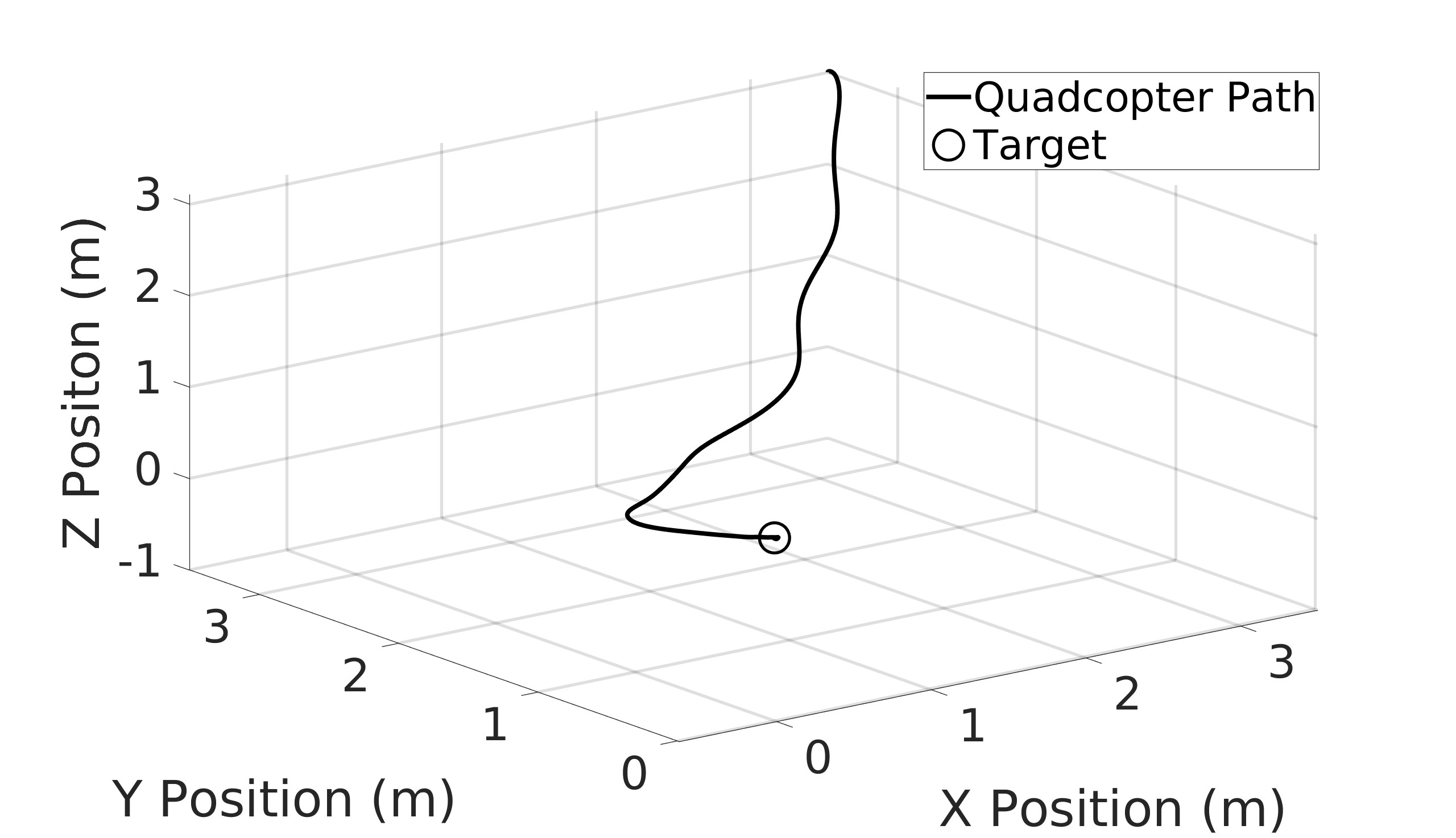}
            \caption{Response of stochastic agent in large environment from an initial position of [3.5, 3.5, 3]}
            \label{offS4.5}
    \end{figure}

   Fig. \ref{offS4.5} shows the response of the stochastic training agent when starting with an initial position of [3.5, 3.5, 3]. The stochastic training agent successfully stabilized at the target position as shown although the agent was never trained on this point previously.
   
    Not only did the stochastic agent learn the entirety of the environment, but also the stochastic agent successfully reached the target when presented with an initial position outside the trained environment.

    In contrast to deterministic algorithms, stochastic algorithms train a sample-efficient policy where each observation has a probability distribution over actions to maximize the potential reward. By adding an entropy term to the policy parameters along with the reward, stochastic algorithms encourage exploration while maximizing the rewards promoting a balance between exploration and exploitation leading to a more stable algorithm \cite{haarnoja2018soft}.

    Higher entropy results in more stochasticity, preventing the policy from collapsing into determinism. Dynamic entropy tuning in stochastic algorithms optimizes the entropy coefficient online, allowing the algorithm to learn when to explore and when to exploit maximizing the exploration-exploitation trade-off.

\section{Conclusion}
In this paper, the effect of dynamic entropy tuning in stochastic RL algorithms on controlling and stabilizing the quadcopter was presented. The exploration efficiency of dynamic entropy tuning is compared with the traditional approach of adding external noise in deterministic algorithms. The TD3 and SAC algorithms were chosen to represent the deterministic and stochastic algorithms respectively. The training results demonstrated that the stochastic algorithm with dynamic entropy tuning outperformed the deterministic algorithm regarding learning speed and efficiency within the provided environment. The stochastic algorithm achieved higher rewards and had a more stable learning process. Training results also showcased that adding external noise allows catastrophic forgetting to occur leading to a fast decay in performance and rewards. In addition to this, the simulation results showed the ability of stochastic algorithms to generalize its policy and its ability to control the quadcopter in unseen states. The dynamic entropy tuning approach not only increases learning efficiency and stability but also prevents the risks of catastrophic forgetting, ensuring stable performance and adaptability in quadcopter control.

\bibliographystyle{ieeetr}
\bibliography{references}
\end{document}